\documentclass[11pt,a4paper]{article}
\usepackage{geometry}
\usepackage{subcaption}

\usepackage{interval}
\usepackage{array, makecell} 
\usepackage{algorithm}
\usepackage{booktabs}
\usepackage{mathtools, amsmath, amsthm, amssymb}
\usepackage{bm} % bold math
\usepackage{bbm} % blackboard math \mathbbm

\newcommand{\Z}{ \ensuremath{\mathbb{Z}}}  % integers

\newcommand{\R}{ \ensuremath{\mathbb{R}}}  % real numbers
\DeclarePairedDelimiterX\set[1]{\lbrace}{\rbrace}{  #1 }  % set* autoscales
\DeclarePairedDelimiterX\norm[1]{\lVert}{\rVert}{#1}  			% norm (\norm*{} autoscales or \norm[\big]{})
\DeclarePairedDelimiterX\inner[2]{\langle}{\rangle}{#1 \,,\, #2}  	% inner product
\newcommand{\mat}[1]{\mathbf{#1}}

\DeclarePairedDelimiterX\abs[1]{\lvert}{\rvert}{#1} % absolute value: \abs{} = no-resize, \abs*{} = left/right auto-resize, \abs[size-cmd]{} = size manually adjusted with size-cmd = \big,\Big,\bigg,\Bigg

\title{Predicting the Critical Number of Layers for Hierarchical Support Vector Regression\thanks{This research was funded by DARPA grant number HR0011-18-9-0033.}}

\author{Ryan Mohr\thanks{AIMdyn Inc.; mohrr@aimdyn.com}, Maria Fonoberova\thanks{AIMdyn Inc.; mfonoberova@aimdyn.com}, Zlatko Drma\v{c}\thanks{University of Zagreb, Croatia; drmac@math.hr}, Iva Manojlovi\'{c}\thanks{AIMdyn Inc.; imanojlovic@aimdyn.com}, and Igor Mezi\'{c}\thanks{University of California, Santa Barbara; mezic@ucsb.edu}}

\begin{document}	
\maketitle
\begin{abstract}
	Hierarchical support vector regression (HSVR) models a function from data as a linear combination of SVR models at a range of scales, starting at a coarse scale and moving to finer scales as the hierarchy continues. In the original formulation of HSVR, there were no rules for choosing the depth of the model. In this paper, we observe in a number of models a phase transition in the training error---the error remains relatively constant as layers are added, until a critical scale is passed, at which point the training error drops close to zero and remains nearly constant for added layers. We introduce a method to predict this critical scale a priori with the prediction based on the support of either a Fourier transform of the data or the Dynamic Mode Decomposition (DMD) spectrum. This allows us to determine the required number of layers prior to training any models.
\end{abstract}


%%%%%%%%%%%%%%%%%%%%%%%%%%%%%%%%%%%%%%%%%%

\section{Introduction}
Many of the machine learning algorithms require the correct choice of hyperparameters to give the best description of the given data. One of the most popular methods for choosing hyperparameters is doing grid-search. In grid search, the model is trained for some points in hyperparameter space which are called a grid and then the model with the lowest error on the validation set is chosen. Other methods use alternative optimization algorithms, such as genetic algorithms. All these methods include training model and calculating error multiple times, which can be very expensive if the hyperparameter space is large. 

The purpose of this paper is to develop a method for determining hyperparameters for Support Vector Regression (SVR) with Gaussian kernels from time-series data only. Unlike other approaches for choosing hyperparameters, our method identifies a set of hyperparameters without needing to train models and perform grid search (or executing some other hyperparameter optimization algorithm), thereby bypassing a potentially costly step. The proposed method identifies the inherent scale and complexity of the data and adapts the SVR accordingly. In particular, we give a method for determining the scale of Gaussian kernel by connecting it to the most important frequencies of the signal, as determined by either a Fourier transform or Dynamic Mode Decomposition. Thus we leverage classical and generalized harmonic analysis to inform the choice of hyperparameters in modern machine learning algorithms. A pertinent question is why not just use FFT? The answer is that HSVR are better suited to model strongly locally varying data whereas, for FFT to be efficient, the data needs to possess some symmetry such as space or time translation.

\subsection{Previous work} 
There are many different approaches in tuning SVR hyperparameters for reducing generalization error. Some popular methods of estimating generalization error are Leave-one-out (LOO) score and $k$ cross-validation score. They are easy to implement, but for calculating those measures more models have to be trained for each combination of hyperparameters that need to be tested. This can be prohibitively expensive, so other error estimates, which are easier to calculate, were developed. These methods include the Xi-Alpha bound \cite{joachims2003maximum}, the generalized approximate cross-validation \cite{wahba1999support}, the approximate span bound \cite{vapnik2000bounds}, the VC bound \cite{vapnik2000bounds}, the radius-margin bound \cite{vapnik2000bounds} or the quality functional of the kernel \cite{ong2005learning}.

For choosing hyperparameters, the simplest method is to perform grid search over the space of hyperparameters and then choose the ones with the lowest error estimation. However, grid search suffers from the curse of dimensionality, scaling exponentially with the dimension of the hyperparameter configuration space. Other work in hyperparameter optimization (HPO) seek to mitigate this problem. Random search samples the configuration space and can serve as a good baseline \cite{HPO-book:2019, HPO-review:2020}. Bayesian optimization techniques to find optimal hyperparameters, often using Gaussian processes as surrogate functions, offers a more more computational efficient algorithm than grid search or random search requiring fewer attempts to find the optimal parameters \cite{HPO-review:2020}. However, Bayesian optimization in this form requires more computational resources \cite{HPO-review:2020}.

There are also gradient-based approaches \cite{chapelle2002choosing}, \cite{chung2003radius}, \cite{gold2003model}.
There are also several derivative-free optimization methods. For example, in \cite{hooke1961direct}, a pattern search methodology for hyperparameters, the parameter optimization method based on simulated annealing \cite{kirkpatrick1983optimization}, Bayesian method based on MCMC for estimating kernel parameters \cite{mallick2005bayesian}, and also methods based on Genetic algorithms \cite{friedrichs2005evolutionary}, \cite{frohlich2003feature}, \cite{igel2005multi}. 

What all of these methods have in common is that they require training a model and evaluating the error for each parameter set that needs to be tried. This can be quite expensive, both in time and computational resources, if there is a lot of data or we want to train multiple models at the same time. They are iterative, which means that we usually do not know how many models will be trained before getting an estimation of the best hyperparameters. There are methods, such as early cutoff, that try to reduce these burdens, but fundamentally a model needs to be trained for each set of candidate hyperparameters.

In contrast, our approach gives a set of hypeparameters without ever computing a single model. Specifically, in this paper, we are interested in modeling multiscale signals with a linear combination of SVRs. Two fundamental questions are (1) how many SVR models are going to be used? and (2) what should the scale be for each SVR model. Clearly the dimension of the configuration space for the scale parameters is conditional on the number of layers. Our methods are built off fast spectral methods like fast Fourier transform and Dynamic Mode Decomposition and return both the number of layers (number of SVR models) and the scales for each SVR without ever having to train a model. Bypassing the expensive step of computing a model for each candidate set of hyperparameters can lead to dramatic savings in computational time and resources.

%%%%%%%%%%%%%%%%%%%%%%%%%%%%%%%%%%%%%%%%%%
\section{Methods}

\subsection{Support Vector Regression}
Support Vector Machines (SVM) have been introduced in \cite{cortes1995support} as a method for classification. The goal was to set hyperplane between classes, where hyperplane is defined by linear combination of a subset of a training set, called Support Vectors. The problem is formulated as a quadratic optimization problem that is convex so it has a unique solution. The problem with SVM is that it can only find a linear boundary between classes which is often not possible. The trick is to map the training data to a higher dimensional space and then use kernel functions to represent the inner product of two data vectors projected onto this space. Then only inner products are needed for finding the best parameters. The advantage of this approach is that we can implicitly map data onto infinite dimensional space. One of the most used kernels is Gaussian function defined by
\begin{equation}\label{eq:sigma-gaussian}
k(x,x') = \exp \left( -\frac{\|x-x'\|_2^2}{\sigma^2}\right) = \exp(-\gamma \|x-x'\|_2^2),
\end{equation}
where $x$ and $x'$ are $n$-dimensional feature vectors, $\sigma^2$ is the variance of the Gaussian function, and $\gamma = 1 / \sigma^2$ is the scale parameter that is usually specified as an input parameter to SVR toolboxes. 

The SVM approach has been extended in \cite{smola2004tutorial} to regression problems and it is called Support Vector Regression (SVR). Let $S = \{(x_1,y_1),\ldots,(x_n,y_n)\}$ be the training set, where $x_i$ is vector in input space $X \subset \mathbf{R}^D$ and $y_i\in \mathbf{R}$ desired output. The aim of SVR is to find a regression function $f : X \rightarrow \mathbf{R}$:
\begin{equation}
f(x) = \omega^T \Phi(x) + b,
\end{equation}
where $\omega$ is the weight vector and $\Phi$ is a mapping of the data points to a higher-dimensional space and $b$ is threshold constant. $\omega$ and $b$ can be found by solving following optimization problem:
\begin{align}
&\min_{\omega,b}    \frac{1}{2}\omega^T\omega + C\Sigma_{i=1}^n E_i^+ + C\Sigma_{i=1}^n E_i^-, \\
&s.t\qquad  y_i - \omega^T\Phi(x) - b \leq \epsilon + E_i^+ \\
&\qquad\quad \omega^T \Phi(x) + b - y_i \leq \epsilon + E_i^- \\
&\qquad\quad E_i^+, E_i^- \geq 0, \qquad (i = 1, \ldots ,n).
\end{align}
The parameter $\epsilon$ determines width of tube around the regression curve and points inside it do not contribute to the loss function. Parameter $C$ adjusts the trade off between the regression error and regularization , $E^+$ and $E^-$ are slack variables for relaxing approximation constraints and measure the distance of each data point from the $\epsilon$ tube. In practice, the dual problem is solved, which can be written as:
\begin{align}
&\max_{\alpha^+,\alpha^-} \quad  -\frac{1}{2}(\alpha^+-\alpha^-)K(\alpha^+-\alpha^-) - \epsilon \sum\limits_{i=1}^n(\alpha_i^+ + \alpha_i^-) + \sum\limits_{i=1}^n y_i(\alpha_i^+ - \alpha_i^-), \\
&s.t \quad \quad \sum\limits_{i=1}^n (\alpha_i^+ - \alpha_i^-) = 0 \\
&\quad \quad \quad \alpha_{i}^+, \alpha_{i}^- \in \interval{0}{C}, \qquad (i = 1, \dots,n).
\end{align}
where $\alpha^+,\alpha^- \in \mathbf{R}^n$ are the dual variables and $K \in \mathbf{R}^{n \times n}$ is the kernel matrix evaluated from a kernel function, $K_{ij} = k(x_i,x_j)$ where $k(x,x')$ is the kernel function. Solving that problem, the regression function becomes:
\begin{equation}
f(x) = \sum\limits_{i=1}^n (\alpha_i^+ - \alpha_i^-)k(x,x_i) + b.
\end{equation}
Coefficients satisfy following conditions: 
\[|\alpha_i^+-\alpha_i^-| = \begin{cases} 
0 & \|y_i - f(x_i)\| < \epsilon\\
\in (0, C) & \|y_i - f(x_i)\| = \epsilon \\
C & \|y_i - f(x_i)\| > \epsilon
\end{cases}
\]
Data points for which $|\alpha_i^+-\alpha_i^-|$ is non-zero are called Support Vectors.

\subsection{Dynamic Mode Decomposition (DMD)}
Dynamic Mode Decomposition(DMD) was introduced in \cite{schmid2010dynamic} as a method for extracting dynamic information from flow fields that are either generated by numerical simulation or measured in physical experiment. Rowley et al connected DMD with Koopman operator theory \cite{rowley2009spectral}.

Let the data be expressed in a series of snapshots, given by matrix $\mathbf{V^N_1}$: 
\begin{equation}
	\mathbf{V^N_0} = \{v_0,v_1, \cdots, v_N\},
\end{equation}
where $v_i \in \R^m$ stands for the i-th snapshot of the flow field. We assume there exists a linear mapping $\mathbf{A}$ which relates each snapshot $v_i$ to next one $v_{i+1}$,
\begin{equation}
	v_{i+1} = \mathbf{A}v_i,
\end{equation}
and that this mapping is approximately same during each sampling interval, so we approximately have
\begin{equation}
\mathbf{V^N_0} = \{ v_0, \mathbf{A} v_0,\cdots, \mathbf{A}^{N}v_0\}.
\end{equation} 
We assume that characteristics of the system can be described by the spectral information in $\mathbf{A}$. This information is extracted in a data-driven manner using $\mathbf{V^N_0}$. The idea is to use $\mathbf{V^N_0}$ to construct an approximation of $\mathbf{A}$. In \cite{schmid2010dynamic}, this was done as follows. Define $\mathbf{X} = \mathbf{V_0^{N-1}}$ and $\mathbf{Y} = \mathbf{V_1^{N}}$. Let the singular value decomposition of $\mathbf{X}$ be $\mathbf{X} = \mat{U}\Sigma\mat{V^*}$. Then the representation of a compression of $\mat{A}$ is defined as
	%% ------------ EQUATION ------------ %%
	\begin{equation}\label{eq:schmid-matrix}
	\tilde{\mat{A}} = \mat{U}^* \mat{Y} \mat V \Sigma^+
	\end{equation}
	%% ---------------- END  ---------------- %%
where $\Sigma^+$ is the Moore-Penrose pseudo-inverse of $\Sigma$. Note that \eqref{eq:schmid-matrix} is analytically equivalent to $\mat{U^* A U}$, the compression of $\mat{A}$ to the subspace spanned by the columns of $\mat X$, if $\mat Y = \mat{AX}$. Eigenvectors and eigenvalues of $\mathbf{A}$ are approximations of eigenvectors and eigenvalues of the Koopman operator. 
 
The Koopman operator is an infinite dimensional, linear operator $K$ that acts on all scalar functions $g$ on $M$ as 
\begin{equation}
	Kg(x) = g(f(x)),
\end{equation}
where $f$ is a dynamical system such that $x_{k+1} = f(x_k)$. Let $\lambda_j$ and $\phi_j$ be eigenvalues and eigenfunctions i.e.
\begin{equation}
	K\phi_j(x) = \lambda_j \phi_j(x).
\end{equation}
Let $g(x)$: $M \rightarrow \mathcal{R}^p$ be vector of any quantities of interest. If $g$ lies in span of $\phi_j$, then it can be written as 
\begin{equation}
	g(x) = \sum_{j=1}^{\infty} \phi _j(x)v_j.
\end{equation} 
Then we can express $g(x_k)$ as 
\begin{equation}
g(x_k) = K^kg(x_0) = K^k \sum_{j=1}^{\infty} \phi_j(x_0)v_j = \sum_{j=1}^{\infty} \lambda_j^k \phi_j(x_0) v_j.
\end{equation}
The Koopman eigenvalues, $\lambda_j$ characterize the temporal behaviour of the corresponding Koopman mode $v_j$, the phase of $\lambda_j$ determines its frequency, and the magnitude determines the growth rate.

\subsection{Hierarchical Support Vector Regression}

\begin{figure}[ht]
	\centering
	\includegraphics[width = 0.65\textwidth]{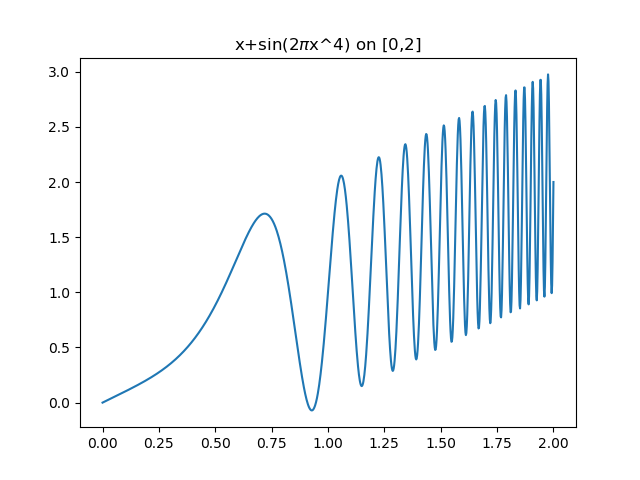}
	\caption{Multiscale example function: $f(x) = x + sin(2\pi x^4)$.}
	\label{example}
\end{figure}

\begin{figure}[ht]
	\centering
	\begin{subfigure}[t]{0.45\textwidth}
		\centering
		\includegraphics[width = 1\textwidth]{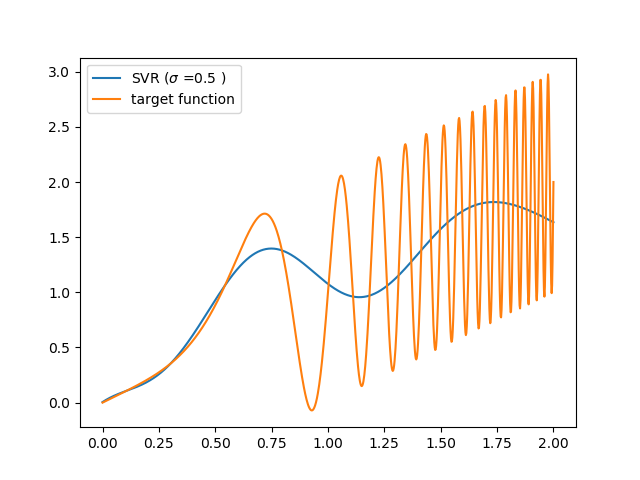}
		\caption{$\sigma=0.5$}
	\end{subfigure}
	\hfil
	\begin{subfigure}[t]{0.45\textwidth}
		\centering
		\includegraphics[width = 1\textwidth]{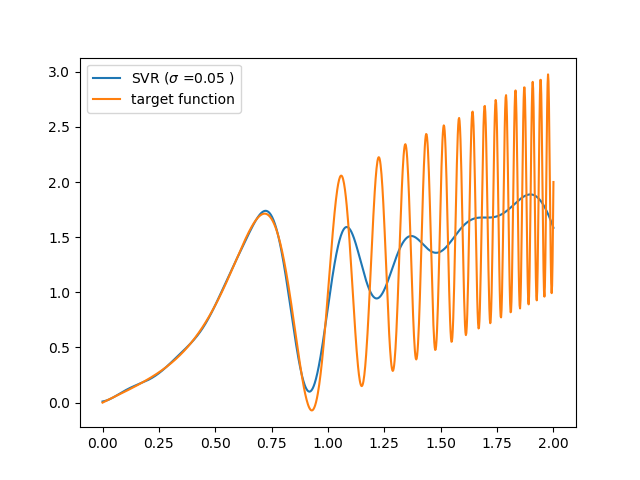}
		\caption{$\sigma=0.05$}
	\end{subfigure}
	\par
	\vspace{0.5cm}
	\begin{subfigure}[t]{0.45\textwidth}
		\centering
		\includegraphics[width = 1\textwidth]{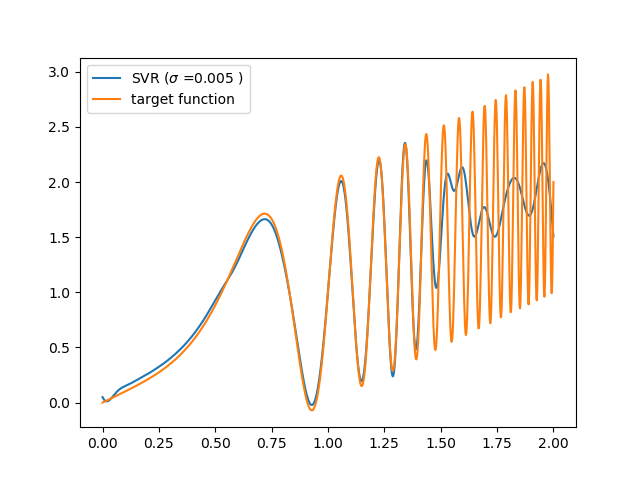}
		\caption{$\sigma=0.005$}
	\end{subfigure}
	\hfil
	\begin{subfigure}[t]{0.45\textwidth}
		\centering
		\includegraphics[width = 1\textwidth]{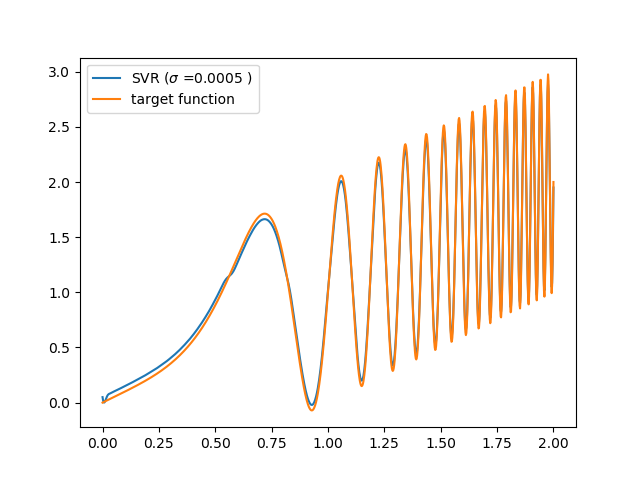}
		\caption{$\sigma=0.0005$}
	\end{subfigure}	
	\caption{SVR results with different scale of Gaussian kernel $\sigma$ = 0.5, 0.05, 0.005, 0.0005. A large kernel provides smooth regression, but cannot reconstruct the details. A small kernel overfits, is unable to generalize, and can be sensitive to noise.}
	\label{svr}
\end{figure}

Classical SVR models which use kernels of a single scale have difficulties approximating multiscale signals. For example, consider the function $f(x) = x + \sin(2\pi x^4)$, for $x\in[0,2]$, whose graph is given in Figure \ref{example}. This function has a continuum of scales. Figure \ref{svr} highlights the difficulties that classical single-scale SVR has in modeling such signals. Using too large a scale $\sigma$, the detailed behavior of the data set is not captured \cite{bellocchio2012hierarchical}; such a model may, however, be useful for a coarse-scale extrapolation outside the training set. Models employing a very small scale $\sigma$ can capture the training set in detail. However, this makes them very sensitive to noise in the training and severely limits the model's ability to generalize outside the training set \cite{bellocchio2012hierarchical}.

\begin{figure}[htb]
\begin{center}
\includegraphics[width=0.75\textwidth]{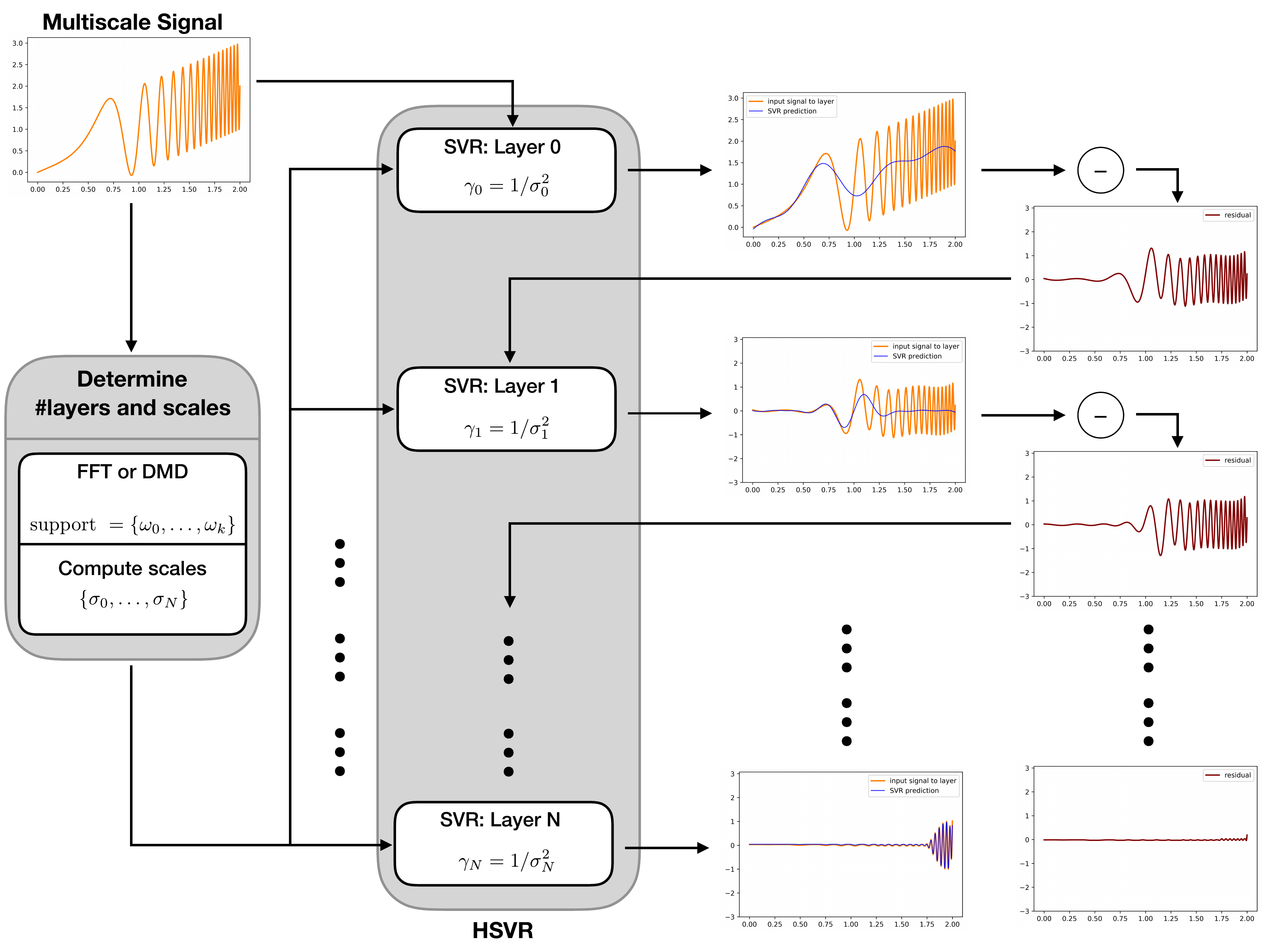}
\caption{Flowchart of the HSVR modeling process. The input data is first used to compute the scales used for the HSVR model (see Alg. \ref{scales}, \ref{hankel}, and \ref{filtering scales}). At layer 0, an SVR model is trained at the coarsest scale $\gamma_0$. The residual is computed by taking the difference between the signal and the model. This residual is then modeled with an SVR model at the next coarsest scale $\gamma_1$. A new residual is computed by taking the difference of the old residual and the $\gamma_1$ SVR model. This process is repeated until the pre-computed scales are exhausted.}
\label{fig:hsvr-diagram}
\end{center}
\end{figure}

In  \cite{bellocchio2012hierarchical}, the authors introduced a multiscale variant of Support Vector Regression which they termed Hierarchical Support Vector Regression (HSVR). The idea behind HSVR is to train multiple SVR models, organized as a hierarchy of layers, each with different scale $\sigma$. The HSVR model is then a sum of those individual models, which we write as
	%% ------------ EQUATION ------------ %%
	\begin{equation}\label{HSVR}
	S(x) = \sum_{\ell=0}^L a_\ell(x; \sigma_\ell, \epsilon),
	\end{equation}
	%% ---------------- END  ---------------- %%
where $L$ is number of layers and $a_\ell(x;\sigma_\ell, \epsilon)$ is SVR model on layer $\ell$ with Gaussian kernel with parameter $\sigma_\ell$. Each SVR layer realizes a reconstruction of the target function at a certain scale. Training the HSVR model precedes from coarser scales to finer scales as follows. Let $\sigma_0 > \sigma_1 > \sigma_L > 0$ be specified. For $\sigma_0$, an SVR model $a_0(x; \sigma_0, \epsilon)$ is trained on the signal $f(x)$ (the 0-th residual) and the residual $r_1(x) = f(x) - a_0(x; \sigma_0, \epsilon)$ is computed. We then proceed inductively for $\ell \geq 1$. Given the residual $r_\ell(x)$, train a model $a_\ell(x; \sigma_\ell, \epsilon)$ to approximate it and compute new residual $r_{\ell + 1}(x) = r_\ell(x) - a_\ell(x; \sigma_\ell, \epsilon)$. The $(L+1)$-th residual is then
	%% ------------ EQUATION ------------ %%
	\begin{equation}
	r_{L+1}(x) = f(x) - a_0(x; \sigma_0, \epsilon) - \cdots - a_L(x; \sigma_L, \epsilon) = f(x) - S(x).
	\end{equation}
	%% ---------------- END  ---------------- %%
Graphically, the process looks like Figure \ref{fig:hsvr-diagram}.

The HSVR model contains a number of hyperparmeters that need to be specified, namely, $\epsilon$, $C_\ell$, the number of layers $L$ to take, and the specific scales $\sigma_\ell$ for those layers. In \cite{bellocchio2012hierarchical}, the authors chose to use exponential decay relationship between the scales, such as $\sigma_{\ell+1} = \sigma_\ell / \sqrt{decay}$. This, however, still leaves the critical choices of $\sigma_0$ and $L$ unspecified. In all of our experiments that follow, we choose $\epsilon$ to be 1 percent of the variation of the signal
	%% ------------ EQUATION ------------ %%
	\begin{equation}
	\epsilon = 0.01( \max f(x) - \min f(x)).
	\end{equation}
	%% ---------------- END  ---------------- %%
For each layer, $C_\ell$ was specified as
	%% ------------ EQUATION ------------ %%
	\begin{equation}
	C_\ell = 5(\max_i r_{\ell-1}(x_i) - \min_i r_{\ell-1}(x_i)).
	\end{equation}
	%% ---------------- END  ---------------- %%
which was the choice given in \cite{bellocchio2012hierarchical}. Additionally, the $decay$ variable was chosen to be 2 so that $\sigma_{\ell+1} = \sigma_{\ell} / \sqrt{decay}$. Python's \textsf{scikit-learn} library \cite{scikit-learn} was used throughout this paper. Its implementation of SVR requires the input parameter $\gamma$, rather than $\sigma$ from \eqref{eq:sigma-gaussian}. These two parameters are related as $\gamma = 1 / \sigma^2$. Thus, the equivalent decay rate of the input parameter is 
	%% ------------ EQUATION ------------ %%
	\begin{equation}
	\gamma_{\ell+1} = \gamma_{\ell} * decay = \gamma_{\ell} * 2. 
	\end{equation}
	%% ---------------- END  ---------------- %%

In the next sections, we take up the task of efficiently determining the hyperparameters $L$ and $\sigma_0$ without the expensive step of performing grid search or training any models.

\section{Predicting the depth of models}
In this section, we will describe how error changes while training HSVR and we will provide methods of estimating the number of layers of such hierarchical model.

\subsection{Phase transition of the training error}
\begin{figure}[htp]
	\centering
	\begin{subfigure}[t]{0.45\textwidth}
		\centering	
		\includegraphics[width=1\textwidth]{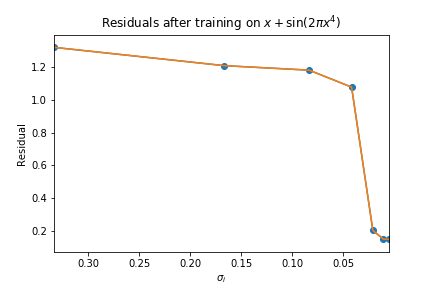}
		\caption{$f(x) = x+\sin(2\pi x^4)$}
	\end{subfigure}
	\quad
	\begin{subfigure}[t]{0.45\textwidth}
		\centering	
		\includegraphics[width=1\textwidth]{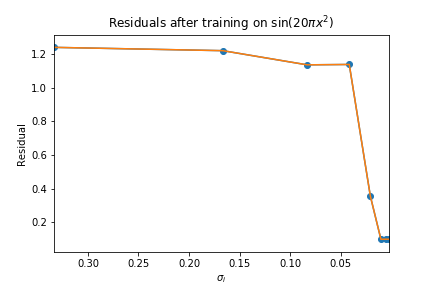}
		\caption{$f(x) = \sin(20\pi x^2 )$}
	\end{subfigure}
	\caption{Residuals while training HSVR model with decreasing $\sigma$ as shown on $x$-axis. Both HSVR models exhibit a phase transition in their approximation error.}
	\label{residuals}
\end{figure}

For the rest of the paper, we have training data set $\set{(x_{train},y_{train})}$ and testing data set $\set{(x_{test},y_{test})}$ and the model $S(x) = \sum_{\ell=1}^L a_\ell(x;\sigma_l)$, as in \eqref{HSVR}. Let $S_i(x) = \sum_{\ell=1}^{i} a_\ell(x;\sigma_l)$ for $i=1,\dots, L$. For each layer, the HSVR (prediction) error is calculated as
	%% ------------ EQUATION ------------ %%
	\begin{equation}\label{errorEqn}
	r_i = \max_{\set{(x_{test}, y_{test})}} \abs{y_{test}-S_{i}(x_{pred})} = \max_{\set{(x_{test}, y_{test})}} \abs{y_{test}-y_{i,pred}},
	\end{equation}
	%% ---------------- END  ---------------- %%
where we have denoted $y_{i,pred} = S_i(x_{pred})$. Therefore, $r_i$ denotes the maximum error between the true signal at the test points $x_{pred}$ and an HSVR model with $i$ layers. By examining the change of the values $r_i$, it can be seen that there is a sudden drop in error, almost until tolerance $\epsilon$, so that adding additional layers cannot reduce error any more. We will call that value the critical-sigma and denote it with $\sigma_c$. The drop in error can be seen on Figure \ref{residuals}.
 
\subsection{Critical scales: intuition and the Fourier transform}
Since the SVR models are fitting the data with Gaussians, a heuristic for choosing the scale will be shown on basic examples of periodic function and then expanded on more general cases in next sections.

\begin{figure}[ht]
	\centering
	\begin{subfigure}[t]{0.45\textwidth}
		\centering
		\includegraphics[width=1\textwidth, height=1.843in]{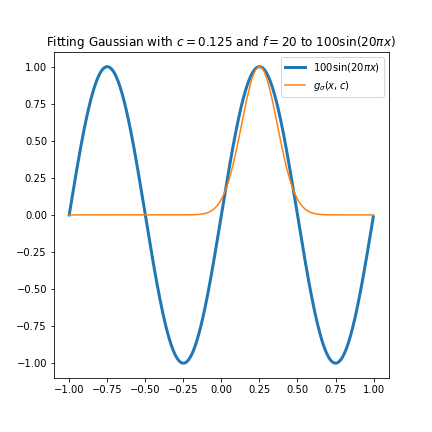}
		\caption{Gaussian with mean 0.25 and standard deviation $\sigma = 1/6$ compared with $\sin(2\pi x)$}
	\end{subfigure}
	\quad
	\begin{subfigure}[t]{0.45\textwidth}
		\centering
		\includegraphics[width=1\textwidth]{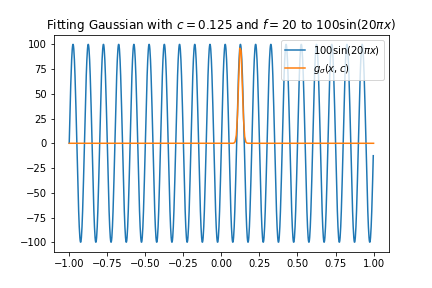}
		\caption{$100$ times a Gaussian with mean 0.125 and deviation $\sigma = 1/20$ compared with $100\sin(20 \pi x)$}
	\end{subfigure}
	\caption{Fitting Gaussians to half-periods of sinusoids using the heuristic \eqref{eq:scale-heuristic}.}
	\label{heuristic1}
\end{figure}

Let $f(x) = \sin(2\pi fx)$, where $x$ is in meters and $f$ is the frequency in cycles per meter. The frequency will be related to the scale, $\sigma$, of the Gaussian: 
\begin{equation}
G_{\sigma}(x,c) = \exp\left(-\frac{\|x-c\|_{2}^2}{\sigma^2}\right).
\end{equation}
To relate the maximum frequency, $f$, to the scale, $\sigma$, we use the heuristic that we want $3$ standard deviations of the gaussian to be half of the period. That is we want
\begin{equation}
3\sigma = \frac{T}{2},
\end{equation}
Since $T= \frac{1}{f}$, then 
\begin{equation}\label{eq:scale-heuristic}
3\sigma = \frac{1}{2f}.
\end{equation}
Figure \ref{heuristic1} justifies this heuristic.

\subsection{Determining scales with FFT}
We assume that we can learn HSVR model with scales $\sigma$ corresponding to important frequencies in the FFT of the signal. The assumption is that these will give an information how to train the model. If in the FFT of the signal, there are a lot of frequencies, we assume that we need more HSVR layers and each will learn the most dominant scale at this level.

Our assumption is that required number of scales in the HSVR model can be determined by the data. If there are a lot of frequencies in FFT of the signal related to relatively big coefficients, we can assume that signal is more challenging for single SVR to model it. As in \cite{bellocchio2012hierarchical}, we refine the scales. Here we use exponential decay and use only frequencies for which corresponding coefficients in FFT are large enough in order to avoid numerical problems and adding insignificant frequencies. The procedure is summarized in Algorithm \ref{scales}. The \emph{Filtering scales} algorithm is summarized in Algorithm \ref{filtering scales}.

\begin{algorithm}[!ht]
	%\SetAlgoLined
	\caption{Determining scales of HSVR model}

		\textbf{Input:} $(x_i,y_i),i=0,\cdots n-1$, where $x_i$ are equidistant points in domain and $y_i$ values of function we want to model  \\ \\
		1: $dx = x[1]-x[0]$ \\
		2: $freq=$ FFT frequencies of the signal \\
		3: $C = FFT(y)$ \\
		4: $C = C/\max(|C|)$ \# normalize coefficients respect to $L1$-norm \\
		5: $freq_{support} = freq[|C| > 0.01]$ \\
		6: $scales = dx / (6 * freq_{support})$ \\
		7: $scales$ = sort scales in descending order \\
		8: $scales = filter(scales)$ \\
		9: \textbf{return} $scales$ 
	\label{scales}
\end{algorithm}

\begin{algorithm}[!ht]
	%\SetAlgoLined
	\caption{Filtering scales}
	\textbf{Input:} \\ $scales$ = vector of scales determined from FFT, \\
	$decay$\\ 
	
	1: $scales_{filtered} = [scales[0]]$ \\
	2: $n = len(scales)$ \\
	3: \textbf{for} i in range(1,n): \\
	4: \quad \textbf{if} $scales_{filtered}[-1] / scales[i] >= decay$: \\
	5: \qquad $scales_{filtered}.append(scales[i])$ \\
	6: \textbf{return:} $scales_{filtered}$
\label{filtering scales}
\end{algorithm}

\begin{algorithm}
	\caption{Train HSVR}
	\textbf{Input:} $(x_i,y_i), i = 0, \cdots n-1$ \\ 
	scales (output of Algorithm \ref{scales}) \\ 
	
	1: $\epsilon = 0.01(\max_{i}(y_i)-\min_{i}(y_i))$ \\ 
	2: $r_0 = y = [y_0, \dots, y_{n-1}]$ \\ 
	3: model = $[\,\, ]$  \# comment: empty list to hold the SVR model at each layer\\
	4: $m = len(scales)$  \# comment: number of HSVR layers \\
	5: \textbf{for} i in range(0, m): \\
	6: \quad	$\sigma_i$ = scales[i] \\
	7: \quad 	$C_i = 5(\max(r_i)-\min(r_i))$ \\
	8: \quad	$svr_i$ = fitted SVR on $(x,r_i)$ with parameters $\sigma_i$, $C_i$ and tolerance $\epsilon$ \\ 
	9: \quad   predictions = $svr_i$.predict(x) \\
	10: \quad   $r_{i+1} = r_{i} - predictions$ \\
	11: \quad model.append($svr_i$) \\ 
	12: \textbf{return:} model
	\label{hsvr training}
\end{algorithm}

\subsection{Determining scales with Dynamic mode Decomposition}
Let  $(x_i, y_i)$, $i=0, \dots, n$  be training set. Output data $y_i$   is  organized into Hankel matrix $Y$ with M rows and N columns  $(M>N)$. We will extract relevant frequencies with Hankel DMD, described in \cite{arbabi2017ergodic}, using the DMD\_RRR described in \cite{Drmac2018data}. The DMD\_RRR algorithm returns the residuals (rez) which determine how accurately the eigenvalues are computed, the eigenvalues ($\lambda$), and the eigenvectors (Vtn). As before, suppose we have a signal 
\begin{equation}
f(x_n) = f(n \delta x), \quad (n = 0, \ldots , N - 1).
\end{equation}
We map this scalar-valued functions into a higher-dimensional space by delay-embedding. We choose $ M < N$. The delay-embedding of the signal is the matrix 
	%% ------------ EQUATION ------------ %%
	\begin{equation}
	H = \begin{bmatrix}
	f(x_{0})      & 	f(x_{1}) & \dots   & 	f(x_{N-M})   \\
	f(x_{1})      & 	f(x_{2}) & \dots   & 	f(x_{N-M-1}) \\
	\vdots      & \vdots   & \ddots  & \vdots   \\
	f(x_{M-1})  & f(x_{M})   &\dots     & 	f(x_{N-1})
\end{bmatrix},\label{eq:gen-hankel-matrix}
	\end{equation}
	%% ---------------- END  ---------------- %%
so that for $ j = 0, \ldots, N-M$ 
\begin{equation}
H[:,j] = \begin{bmatrix}
f(x_j) \\ f(x_{j+1}) \\ \vdots \\ f(x_{M+j-1}).
\end{bmatrix}
\end{equation}
We define a generalized Hankel matrix as an $ m\times n$ rectangular matrix whose entries $H_{i,j}$ satisfy
	%% ------------ EQUATION ------------ %%
	\begin{equation}
	H_{i,j} = H_{i+k, j-k}
	\end{equation}
	%% ---------------- END  ---------------- %%
for all indices such that $0 \leq i, i + k \leq m-1$ and $0 \leq j, j-k \leq n-1$, where $k \in \Z$. In simpler terms, a generalized Hankel matrix is a rectangular matrix that is constant on anti-diagonals. A generalized Hankel matrix can also be thought of a submatrix of a larger, regular Hankel matrix. Clearly, the delay-embedding of the scalar signal, Eq.\,\eqref{eq:gen-hankel-matrix}, is an example of a generalized Hankel matrix which is why it was denoted as $H$.

The input matrices for DMD algorithms will be $\mathbf{X}$ and $ \mathbf{Y}$, where $\mathbf{X}$ is the first $N-M$ columns of $\mathbf{H}$ and $\mathbf{Y}$ is the last $N-M$ columns.
Frequencies are calculated as follows:
\begin{equation}
\omega_i = \frac{1}{2 \pi i} \ln\left(\frac{\lambda_i}{ | \lambda_i |} \right);
\end{equation}
i.e., we just scale the DMD eigenvalues so that they are on the unit circle and then extract frequency of the resulting complex exponential, $\exp(i 2\pi (f_\lambda \Delta x)) = \lambda / \lvert \lambda \rvert$. Let $\Omega$ contain the values $\omega_\lambda = f_\lambda \Delta x$. These are directly analogous to the frequencies computed using FFT.

We replace the support of the FFT with the support of the DMD frequency as follows. We will take all the values of $\Omega$ whose corresponding residual is less than some specified tolerance, \texttt{tol}, \emph{and} whose corresponding mode's norm is greater than some percentage, $\eta$, of the total power of the modes. In other words, we only consider the frequencies which were calculated accurately enough and whose modes give a significant contribution to the signal; $\|mode[j]\|$ is analogous to the modulus of a FFT coefficient.

For each mode $Vtn[:,i]$ the energy is $c_i$ for which $\|Y[:,0]-c*Vtn[:,i]\|_2$ is minimal. Total power is defined as 
\begin{equation}
T = \left(\sum \limits_{v_i } c_i^2\right)^{\frac{1}{2}}.
\end{equation}
Like in determining $\sigma_c$ using FFT, the frequency support of $S_{DMD} $ is defined as
\begin{equation}
S_{DMD} = \{ \omega_i : rez[i] < tol, |c_i| > \eta T \},
\end{equation}
where $rez[i]$ is residual corresponding to the $i$-th mode. These considerations can be summarized in Algorithm \ref{hankel}.

\begin{algorithm}
	\caption{Estimating scales from data using Hankel DMD} 
	\label{Hankel}
		 \textbf{Input:} \\ 
		 time step: $\Delta x$, \\
		 time series f: $f[n] = f(\Delta x n)$, \\
		 length of time series vector: $N$, \\
		 tolerance for support:$tol$ , $\eta$, M: number of rows of Hankel matrix \\ 
		 
		 1: H = Hankel matrix made from f with M rows and N-M columns \\
		 2: $rez,\lambda,Vtn = DMD\_RRR(H)$ \\
		 3: $\omega = \frac{1}{2\pi i} \ln\left(\frac{\lambda}{|\lambda|}\right)$ \\
		 4: $T = 0$  \\
		 5: \textbf{for} i = 0 to N - 1: \\ 
		 6: \quad $E[i] = | \langle Y[:,0],Vtn[:,i] \rangle |$ \\
		7: \quad $T = T + E[i]^2$ \\
		8: $T = \sqrt{T}$ \\
		9: $S_{DMD} = []$ \\
		10:  \textbf{for} i=0 to N-1: \\
		11:  \quad \textbf{if} {$rez[i] < tol$ \textrm{ and } $energy[i] > \eta T$} \\
		12: \qquad  $S_{DMD}.append(\omega[i])$  \\
		13: \textbf{return} $\sigma_{DMD}  = \frac{\Delta x}{6 S_{DMD}}$  
		\label{hankel}
\end{algorithm}

%%%%%%%%%%%%%%%%%%%%%%%%%%%%%%%%%%%%%%%%%%
\section{Results}

In this section, results of our methods are provided. We demonstrate our methods on explicitly defined function, system of ODEs, and finally on vorticity data from a fluid mechanics simulations. In all cases,
	%% ------------ EQUATION ------------ %%
	\begin{equation}\label{eq:epsilon-parameter}
	\epsilon = 0.01 \left(\max_{\set{y_{train}}} y_{train} - \min_{\set{y_{train}}} y_{train} \right)
	\end{equation}
	%% ---------------- END  ---------------- %%
and the error is calculated as in \eqref{errorEqn}.

\subsection{Explicitly defined functions}
We model explicitly defined functions $f(x)$ on $[0, 2]$. The dataset consists of $2001$ equidistant points in that interval, where every other point is used for the training set. The error is calculated as the maximum absolute value of difference between prediction and actual value (Equation \eqref{errorEqn}). Results are summarized in Table \ref{exact}. The parameter $\epsilon$ is specified as above. Each method predicts a certain number of layers required to push the model error close to $\epsilon$. As seen in the table, both the FFT and DMD approaches produce models that give similar errors (near $\epsilon$). Although the DMD approach results in models with less layers in general, it fails for polynomials and the exponential function. This is because we normalize eigenvalues to the unit circle. An extension of this could use the modulus $\abs{\lambda}$ as well as $\omega$ in $e^{\lambda+i\omega}$.

\begin{table}[ht]
	\caption{Results for explicitly defined functions, when using scales determined from FFT with decay $2$ and $\epsilon$ given by \eqref{eq:epsilon-parameter}, for $e^x$ DMD did not output frequencies different from $0$.}
	\centering
 \begin{tabular}{cccccc}
	\toprule
	function & $\epsilon$ & \makecell{Predicted \#\\ of layers\\(FFT)} & error(FFT) & \makecell{Predicted \# \\of layers\\(DMD)} & error(DMD) \\ 
	\midrule
	$\sin(2\pi x)$  &0.02 &1 & 0.02 &1  & 0.02 \\
	$\sin(20\pi x)$ &0.0199 & 1   &0.021 &1 & 0.02  \\
	$\sin(200\pi x)$ & 0.019 &1  & 0.093 & 1 & 0.097 \\
	$100\sin(20\pi x)$ & 1.99 &1	  & 2 & 1&2.01  \\
	$40cos(2 \pi x)$ &0.8 &1  &0.8 & 1 & 0.8 \\
	$100cos(20 \pi x)$ &2 &1  &2.03 &1 &2\\
	$sin(2 \pi x^2)$ &0.0199 & 5 & 0.02 &1 &0.02  \\
	$x + x ^2 + x^3$ &0.14 &2  &0.14 &1 & 8 \\
	$e^x$ &0.063 &1  &0.064 & *& *\\
	$x+\sin(2 \pi x^4)$ & 0.03 & 7  & 0.037 &1 &0.034\\
	$\cos(2 \pi x) + \sin(20 \pi x)$ & 0.0397  & 2  & 0.0404 &2 &0.042\\
	$\cos(20 \pi x) \sin(15 \pi x)$ & 0.02 & 2  & 0.021 & 2 &0.022\\
	$\cos(32 \pi x)^3$ & 0.0199 & 1  & 0.022 & 2& 0.022\\
	\hline
\makecell{	$\sin(13 \pi x) + \sin(17 \pi x) +$\\$ \sin(19 \pi x) + \sin(23 \pi x)$} & 0.076 & 1  & 0.077 & 1&0.077 \\	
	\hline
	$ \sin(50 \pi x) \sin(20 \pi x) \cos(15 \pi x)$ &0.0187 &3  & 0.02 & 2 &0.02\\
	\hline
	\makecell{$\sin(40 \pi x) \cos(10 \pi x) +$\\$ 3\sin(20 x)\sin(40x)$} &0.064 &5  & 0.065 &3 & 0.066\\
	\hline
	$\sin(2x)\cos(32x)$ & 0.0198 & 5   & 0.02 & 1&0.02 \\
	\bottomrule
\end{tabular}
\label{exact}
\end{table}

\subsection{ODEs}
We will demonstrate the methods on modeling solutions of the Lorenz system of ODE

%\begin{ceqn}
	\begin{align}
	x(t)& = -10x + 10y, \\
	y(t)& = 28x - y -xz, \\
	z(t)& = xy - \frac{8}{3}z, 
	\end{align}
%\end{ceqn}
with initial conditions $x_1(0) = 1.0$, $x_2(t) = 1.0$, $x_3(t) = 1.0$, on $\interval{0}{10}$. The system evolved solved for $500$ equidistant time steps which was used for training set. A solution consisting of $2000$ equidistant time steps is then used for test set. Each $x(t)$, $y(t)$, $z(t)$ were regarded as separate signals to model. All hyperparameters are determined as before. Results are summarized in Table \ref{odeLorenz}. For both systems, we can see that error drops nearly to $\epsilon$, but the DMD approach results in models with less layers.

\begin{table}[htp]
	\caption{Results of training HSVR with scales used from FFT and DMD on Lorenz system.}
	\centering
	\begin{tabular}{cccccc} 
	\toprule
		function &$\epsilon$ & \makecell{Predicted \# \\ of layers \\ (FFT)} &error (FFT) & \makecell{Predicted \#\\ of layers \\(DMD)} & error(DMD) \\ 
		\midrule
		x(t) & 0.314 & 6&  0.325& 2 & 0.324 \\
		y(t) & 0.408 & 6& 0.469& 2 & 0.469 \\
		z(t) & 0.468 & 5& 0.485 & 2  &0.494  \\
		\bottomrule
	\end{tabular}
	\label{odeLorenz}
\end{table}

\subsection{Vorticity data}
This data was provided by Georgia Institute of Technology \cite{Tithof-2017JFM-Bif-Q2D-Kolmogorov, Suri-2014PhFl-Velocity-Kolmogorov-flow}. It contains information about vorticity of a fluid field is some computational box. For each point in space, we want to model vorticity at that point as time evolves. Vorticity in every other time step is used for training and we test our models on rest of the time steps. We analyzed doubly periodic and non periodic data.

\subsubsection{Doubly periodic data}
\begin{figure}[htp]
	\includegraphics[width=0.3\textwidth]{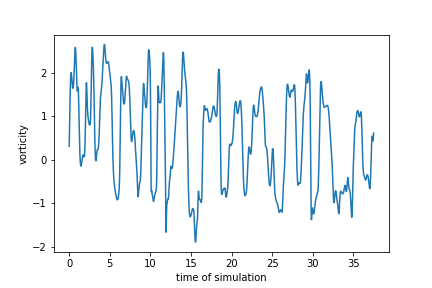}
		\includegraphics[width=0.3\textwidth]{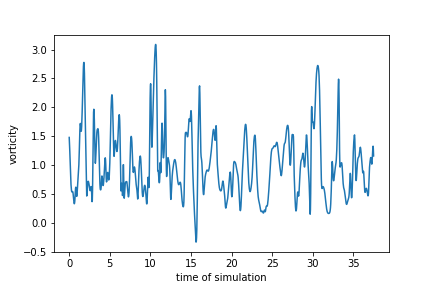}
				\includegraphics[width=0.3\textwidth]{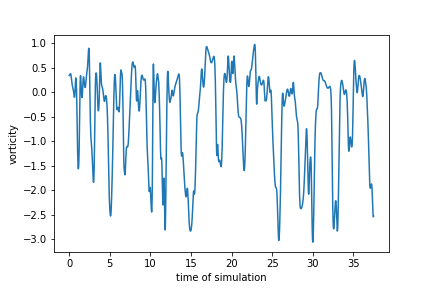}
		\caption{Examples of vorticity at 3 different space-points for fluid simulations with doubly periodic boundary conditions.}
		\label{doublyExamples}
\end{figure}

This data was generated assuming doubly periodic boundary conditions. The dataset contains information about the vorticity on $128 \times 128$ equidistant space-grid on $\interval{0}{0.1016}^2$. There are in total $1201$ snapshots with time step $dt=0.03125$. This results in a tensor of dimensions $128 \times 128 \times 1201$. For each fixed point in space, there is a signal with $1201$ time steps. A few examples of such signals are shown in Figure \ref{doublyExamples}.

\begin{figure}[htp]
	\centering
	\begin{subfigure}[t]{0.45\textwidth}
		\centering
		\includegraphics[width = 1\textwidth]{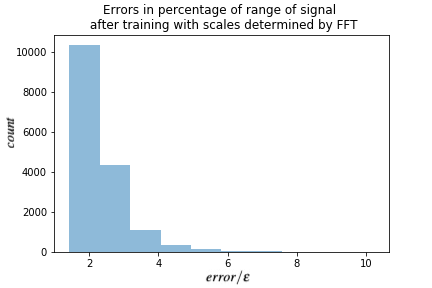}
		\caption{Scales determined by FFT}
	\end{subfigure}
	\quad
	\begin{subfigure}[t]{0.45\textwidth}
		\centering
		\includegraphics[width = 1\textwidth]{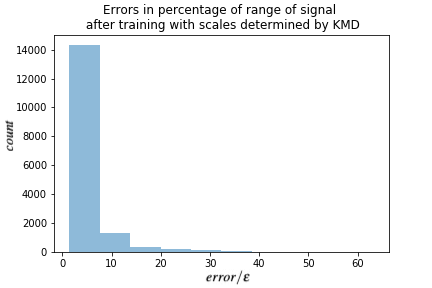}
		\caption{Scales determined by DMD}
	\end{subfigure}
	\caption{Histograms of $error / \epsilon$ for models trained on vorticity data with doubly periodic boundary conditions. $error$ is the model error given by \eqref{errorEqn} (with $i=L$) and $\epsilon$ is given by \eqref{eq:epsilon-parameter}. There were $128^2 = 16,384$ total models trained. The count on the vertical axis is the number of models that fell into the corresponding bin.}
	\label{errorsRelD}
\end{figure}

For each of these signals, every other point in time is used for training the HSVR model, which results in total of $128 \times 128$ models. Results are summarized in Table \ref{doubly}. For each model, the error is calculated as in \eqref{errorEqn} with $i=L$, where $L$ is number of layers. Since $\epsilon$ is given by \eqref{eq:epsilon-parameter}, i.e., 1\% of the range of the training data, the ratio error/$\epsilon$ gives a measure of error in percentages of range of signal. A ratio of 2 would imply that the maximum error of the model over the test set was only 2\% of the range of the test data; i.e. 
	%% ------------ EQUATION ------------ %%
	\begin{equation}
	2 = \frac{error}{\epsilon} \iff error = 2\epsilon = 0.02 \left(\max_{\set{y_{train}}} y_{train} - \min_{\set{y_{train}}} y_{train} \right).
	\end{equation}
	%% ---------------- END  ---------------- %%
Histograms of these ratios for both FFT and DMD are shown on Figure \ref{errorsRelD}. Most models produced on error close the $\epsilon$ threshold (a perfect match would give a ratio of 1).

\begin{table}[htp]
	\caption{Results after training HSVR on doubly periodic vorticity data}
	\centering
	\begin{tabular}{cccccc}
	\toprule
		 &  $\epsilon$ & \makecell{Predicted \#\\of layers\\ (FFT)} & \makecell{error \\(FFT)}  &  \makecell{Predicted \#\\ of layers \\(DMD)}& \makecell{error \\(DMD)} \\ \midrule
		 min & 0.0199 &  6&0.038  &1 &0.0399\\ 
		 mean &0.0354 & 8&0.082 &3 &0.148 \\
		 max &0.0488 & 9&0.284  &5 &2.463 \\
		 \bottomrule
	\end{tabular}
	
	\label{doubly}
\end{table}

\subsubsection{Non periodic data}
\begin{figure}[htbp]
		\includegraphics[width=0.3\textwidth]{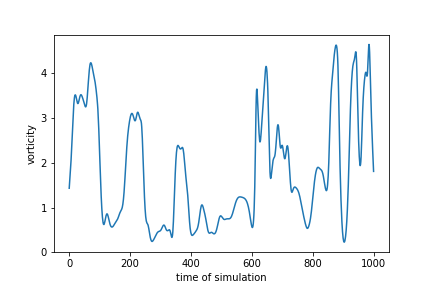}
				\includegraphics[width=0.3\textwidth]{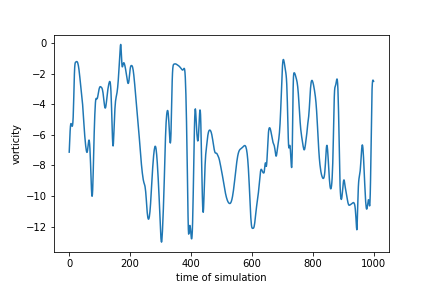}
						\includegraphics[width=0.3\textwidth]{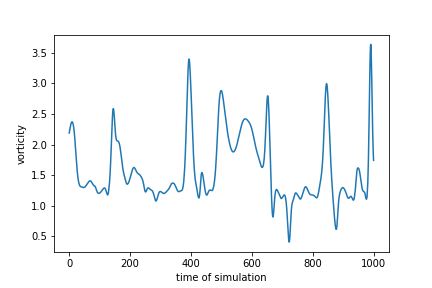}
						\caption{Examples of vorticity in 3 space-points for non periodic data}
						\label{nonperiodicExamples}
\end{figure}

\begin{figure}[ht]
	\begin{subfigure}[t]{0.45\textwidth}
		\centering
		\includegraphics[width = 1\textwidth]{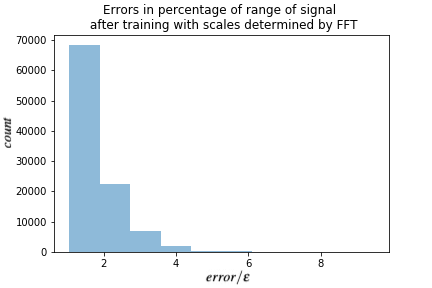}
		\caption{Scales determined by FFT}
	\end{subfigure}
	\quad
	\begin{subfigure}[t]{0.45\textwidth}
		\centering
		\includegraphics[width = 1\textwidth]{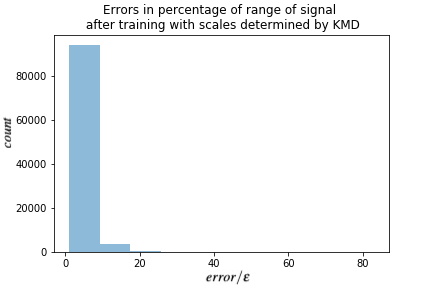}
		\caption{Scales determined by DMD}
	\end{subfigure}
		\caption{Histograms of $error / \epsilon$ for models trained on vorticity data with non-periodic boundary conditions. $error$ is the model error given by \eqref{errorEqn} (with $i=L$) and $\epsilon$ is given by \eqref{eq:epsilon-parameter}. There were $359 \times 279 = 100,161$ total models trained. The count on the vertical axis is the number of models that fell into the corresponding bin.}
	\label{errorsRelNP}
\end{figure}

The data contains information about vorticity on $359 \times 279$ equidistant space-grid on with step $dx = dy = 0.05$. No assumption of periodicity of boundary condition was made. There are in total $1000$ snapshots with time step $dt = 1ms$. This results in tensor $359 \times 279 \times 1000$. Similar to the dataset with periodic boundary conditions, the HSVR model is trained for each fixed point in space, which results in $359 \times 279$ models. The error's and $\epsilon$'s are calculated as before.  A few examples of such signals are in Figure \ref{nonperiodicExamples}. Results are summarized in Table \ref{nonperiodic} and a histogram of ratios error/$\epsilon$ for FFT and DMD are shown in Figure \ref{errorsRelNP}. For both doubly periodic and non periodic data, we can see that a large majority of the models have a ratio between 1 and 2 which means the maximum error of a majority models is $error \leq 2 \epsilon = 0.02 (\max_{\set{y_{train}}} y_{train} - \min_{\set{y_{train}}} y_{train})
$ which is less than 2\% of the range of the training data. For DMD approach, the majority of models have ratios less 10 which corresponds to a maximum error of $error \leq 10 \epsilon = 0.01 (\max_{\set{y_{train}}} y_{train} - \min_{\set{y_{train}}} y_{train})$, which is less than 10\% of the range of the training data. From Tables \ref{doubly} and \ref{nonperiodic} we can see that DMD estimates less layers, but with bigger error after training.

\begin{table}[htbb]
	\caption{Results after training HSVR on non periodic vorticity data}
	\centering
	
	\begin{tabular}{cccccc}
	\toprule
	&  $\epsilon$ & \makecell{Predicted \# \\ of layers\\ (FFT)} & \makecell{error \\(FFT)}  &  \makecell{Predicted \#\\ of layers \\(DMD)}& \makecell{error \\(DMD)} \\
	\midrule
	min & 0.019 & 5 & 0.0006 & 2 & 0.0006  \\ 
	mean &0.035 & 7 & 0.0975& 3&   0.197 \\
	max & 0.048& 9 & 0.667 & 7 &6.137\\
	\bottomrule
\end{tabular}
\label{nonperiodic}
\end{table}

%%%%%%%%%%%%%%%%%%%%%%%%%%%%%%%%%%%%%%%%%%
\section{Discussion}

Both approaches (FFT and DMD) to estimating the number of layers required by HSVR to push the modeling error close to the $\epsilon$ threshold show promise and allow computation of the HSVR depth a priori. While the DMD approach often predicted a smaller number of layers, usually with comparable error, our choice of unit circle normalization prevented it from performing well on functions without oscillation. Furthermore, the DMD algorithm itself comes from the dynamical systems community and requires that the domain of the signal (the $x$ variables) be strictly ordered. For the explicitly defined functions we considered, there was a strict spatial ordering since the domains of the functions were intervals on the real line. For the vorticity data, the time signals at each spatial point were to be modeled and therefore the data points could be strictly ordered in time. For functions whose domain is multi-dimensional, say $\R^2$, there is no strict ordering and the DMD approach, as formulated here, would break down. The method based on the Fourier transform seems to have more promise in analyzing multivariable, multiscale signals, as it can compute multidimensional wave vectors. 

A recent paper \cite{schoenholz2017deep} also deals with the number of layers of a model and “scales”. However, in the mentioned paper, the authors are concerned with how far a signal or gradients will propagate through a network before dying. The scales they compute control how many layers the gradient or signal can propagate before they die. If the network is too deep the gradients go to zero before fully backpropagating through the network, resulting in an untrainable network. The result they compute is a fundamental characteristic of the network and is independent of the dataset. It does not matter what the input data is: constant, single scale, multiscale, etc, the scale parameters they compute are not affected.

Conversely, we are most focused on multiscale signals and tailoring the architecture to best represent such signals. The scales we compute are inherent properties of the dataset, not inherent properties of the network model. The length of the network adapts to the scales contained in the dataset.

It would be interesting in the future to combine the two methodologies, with the methods in \cite{schoenholz2017deep} giving bounds on the number of layers and then adapting our techniques to analyze the inherent scales in the dataset. This could tell us whether a simple, fully, connected, feedforward network could adequately represent the signal or if something more complex was needed.

%%%%%%%%%%%%%%%%%%%%%%%%%%%%%%%%%%%%%%%%%%
\section{Conclusions}
In this work we presented method for choosing hyperparameters for HSVR from time-series data only. We described two approaches, using FFT or DMD. We saw that estimating hyperaparameters with FFT results in model with more layers and smaller error, whereas the DMD approach gave models that had less layers (and thus more efficient models).

%%%%%%%%%%%%%%%%%%%%%%%%%%%%%%%%%%%%%%%%%%
\vspace{6pt} 
\bibliographystyle{plain}
\bibliography{frequency_informed_scale}
%%%%%%%%%%%%%%%%%%%%%%%%%%%%%%%%%%%%%%%%%%
\end{document}